\documentclass[sn-nature,Numbered]{sn-jnl}
\usepackage{graphicx}%
\usepackage{multirow}%
\usepackage{amsmath,amssymb,amsfonts}%
\usepackage{amsthm}%
\usepackage{mathrsfs}%
\usepackage{xcolor}%
\usepackage{textcomp}%
\usepackage{manyfoot}%
\usepackage{booktabs}%
\usepackage{algorithm}%
\usepackage{algorithmicx}%
\usepackage{algpseudocode}%
\usepackage{listings}%
\usepackage{amsmath,amsfonts}
\usepackage{algorithm}
\usepackage{xcolor}
\usepackage{array}
\usepackage{mathtools}
\usepackage{textcomp}
\usepackage{stfloats}
\usepackage{indentfirst}
\usepackage{url}
\usepackage{verbatim}
\usepackage{graphicx}
\usepackage{lettrine}
\usepackage{bm}
\usepackage{mathrsfs}
\usepackage{amsfonts}
\usepackage{multirow}
\usepackage{algpseudocode,algorithmicx}
\usepackage{float}
\usepackage{makecell}
\usepackage{booktabs}

\begin{document}

\title{Binary Feature Mask Optimization for Feature Selection}


\author*[1]{\fnm{Mehmet E.} \sur{Lorasdagi}}\email{efe.lorasdagi@bilkent.edu.tr}
\equalcont{These authors contributed equally to this work.}

\author[1]{\fnm{Mehmet Y.} \sur{Turali}}\email{yigit.turali@ug.bilkent.edu.tr}
\equalcont{These authors contributed equally to this work.}

\author[1]{\fnm{Suleyman S.} \sur{Kozat}}\email{kozat@ee.bilkent.edu.tr}

\affil*[1]{\orgdiv{Department of Electrical and Electronics Engineering}, \orgname{Bilkent University}, \orgaddress{\city{Ankara}, \postcode{06800}, \country{Turkey}}}





\abstract{We investigate feature selection problem for generic machine learning models. We introduce a novel framework that selects features considering the outcomes of the model. Our framework introduces a novel feature masking approach to eliminate the features during the selection process, instead of completely removing them from the dataset. This allows us to use the same machine learning model during feature selection, unlike other feature selection methods where we need to train the machine learning model again as the dataset has different dimensions on each iteration. We obtain the mask operator using the predictions of the machine learning model, which offers a comprehensive view on the subsets of the features essential for the predictive performance of the model. A variety of approaches exist in the feature selection literature. However, to our knowledge, no study has introduced a training-free framework for a generic machine learning model to select features while considering the importance of the feature subsets as a whole, instead of focusing on the individual features. We demonstrate significant performance improvements on the real-life datasets under different settings using LightGBM and Multi-Layer Perceptron as our machine learning models. Our results show that our methods outperform traditional feature selection techniques. Specifically, in experiments with the Residential Building dataset, our General Binary Mask Optimization algorithm has reduced the mean squared error by up to 49\% compared to conventional methods, achieving a mean squared error of 0.0044. The high performance of our General Binary Mask Optimization algorithm stems from its feature masking approach to select features and its flexibility in the number of selected features. The algorithm selects features based on the validation performance of the machine learning model. Hence, the number of selected features is not predetermined and adjusts dynamically to the dataset. Additionally, we openly share the implementation or our code to encourage further research in this area.}


\maketitle
\clearpage
\section*{{Index of Notations and Abbreviations}}

\begin{table}[ht]
\caption{Notations}
\begin{tabular}{ll}
\textbf{Notation} & \textbf{Explanation} \\
\hline
$\bm{X}$ & Feature matrix \\
$\bm{X}'$ & Selected feature matrix \\
$\bm{X}_{\text{new}}$ & Feature matrix from a new, unseen dataset \\
$\bm{m}$ & Binary mask vector \\
$\mathcal{J}$ & Set of indices corresponding to selected features \\
$\bm{\theta}$ & Model parameters \\
$L(\cdot, \cdot)$ & Loss function \\
$f(\cdot, \cdot)$ & Machine learning model \\
$\bm{y}$ & Target variable vector \\
$\bm{y}_{\text{new}}$ & Target variable vector for new, unseen data \\
$N$ & Number of samples in the training split \\
$K$ & Number of samples in the validation split \\
$M$ & Number of features in the dataset \\
$\mathbb{R}^M$ & M-dimensional real vector space \\
$\odot$ & Hadamard product (element-wise multiplication) \\
\end{tabular}
\end{table} 
\begin{table}[ht]
\caption{Abbreviations}
\begin{tabular}{ll}
\textbf{Abbreviation} & \textbf{Definition} \\
\hline
CC & Cross Correlation \\
FLBMO & Fixed-Length Binary Mask Optimization \\
GBMO & General Binary Mask Optimization \\
MI & Mutual Information \\
ML & Machine Learning \\
MLP & Multi-Layer Perceptron \\
MSE & Mean Squared Error \\
RFE & Recursive Feature Elimination" \\
\end{tabular}
\end{table}

\clearpage
\section{Introduction}
The feature selection task is widely studied in the machine learning literature for different applications to mitigate overfitting, which hinders the performance of the model, and increases the computational load, when there is a limited number of training samples or if the underlying data is non-stationary \cite{bishop2006, NSOFS, ghosh2023nonlinear}. We study model-specific feature selection for generic machine learning (ML) models for classification and regression tasks, where we select a subset of the informative features from a given set of features without training the model during feature selection. Note that although model-specific feature selection and feature selection for generic ML models are studied in the literature separately, we introduce a feature selection framework that combines these properties and does not train the ML model during feature selection. The primary objective of this study is to develop a framework for feature selection that uses a suitable approach to identify the most informative feature subsets for generic machine learning models without requiring model retraining during feature selection. Our goals are threefold: to prevent the need for model retraining during feature selection by using an approach that works directly with the original model; to provide a generalized method that can be applied to any machine learning model without requiring intrinsic feature importance attributes, and to show the effectiveness of the provided approach by achieving significant performance improvements across various synthetic and widely used real life datasets and models compared to common approaches.

To this end, we introduce two algorithms for feature selection. The first algorithm performs feature selection by eliminating the features using a feature mask according to how the ML model uses the features in the predictions. Then, the algorithm ends the feature selection if the stopping criterion based on the model performance is reached. In the second algorithm, we eliminate the features in the same way, however, our stopping condition is the number of the selected features, e.g., for applications that can handle a limited number of features due to hardware or application constraints. The feature mask enables us to select features without needing to train the model again, and it selects features according to their importance to the ML model, where the ML model is generic.

In the existing literature, there are three main approaches for feature selection, the wrapper methods \cite{karlupia2023wrapper}, the filter methods \cite{batur2021novel}, and the embedded methods \cite{shi2019feature}. The wrapper methods train the model iteratively with different subsets of features. When a certain criterion is reached, the iterations stop. The wrapper methods select features considering the ML model, but they can be impractical in some scenarios as the ML model needs training after each iteration. For both of our algorithms, we do not train the ML model during the feature selection, unlike the other wrapper methods. Thus, our feature selection algorithms have advantages over the other wrapper methods that train the model many times, e.g., when we have a complex deep learning model that is expensive to train, or the model is pre-trained and we cannot train it but can access its predictions. 

The filter methods select features independent of the ML algorithm, based on the statistical relations of the features with the target variable. The filter methods are computationally light, however, they do not select features considering how the ML model uses the data. Our algorithms consider the performance of the feature subsets in a holistic manner, unlike the filter feature selection methods that use the statistics of each feature independently without considering how the features interact with each other. Thus, our feature selection algorithm chooses the feature subsets based on their interactions with the target in a holistic manner that results in better test performance for the ML models compared to the filter methods. 

The embedded methods perform feature selection within the ML algorithm. The embedded methods are lightweight and select features considering the model, but they are tailored for certain models. Hence, the embedded methods cannot be used for a generic ML model \cite{pudjihartono2022review}. Thus, the embedded methods are highly restrictive as they only work with the ML models they are designed for, whereas our feature selection approaches are compatible with any generic ML model. 

RFE is a wrapper-based feature selection method routinely employed in machine learning \cite{guyon2003}. By systematically removing the least important features, it increases the model efficiency and the prediction accuracy. RFE operates iteratively, fitting an estimator model at each step. The chosen estimator must have an intrinsic feature importance attribute, enabling RFE to discard the least significant feature in each iteration according to the feature importance \cite{guyon2003}. However, RFE needs the ML model to have an intrinsic feature importance attribute. This is a constraint on using RFE for generic ML models. Further, RFE trains the model after eliminating each feature, which is a disadvantage when training the model is expensive or is not possible due to restrictions. Our feature selection algorithms have two advantages over RFE. Firstly, our algorithms do not require the ML model to have an intrinsic feature importance attribute. This enables our algorithms to be suitable for generic ML models, unlike RFE. Secondly, our algorithms do not train the model during feature selection, which is advantageous in certain applications compared to RFE. 

Mutual Information (MI) and Cross Correlation (CC) are filter-based feature selection approaches that provide a metric to measure the statistical interdependence between the variables \cite{vinh2012novel, cover2006}. However, note that CC can capture only the linear relations between the variables, and MI can capture the linear and the non-linear relations. One can use these methods to measure the relationships between each feature and the target variable, then, can select the features that are highly related to the target variable according to these statistical measures.
However, these methods do not consider how the ML model uses the features, i.e., they are unsupervised and oblivious to the underlying task yielding them suboptimal. As an example, some features may be related to the target variable but not useful for the predictive outcomes of the model. Further, the filter approaches do not consider the performance of the feature subset as a whole as they consider the statistics of each feature individually. This can cause the filter approaches to select a less informative subset of features in the end. Our algorithms select features considering the performance of the feature subsets in a holistic way while considering their co-interaction and the performance on the underlying task, i.e., in a supervised and complete manner. Thus, our algorithms can select better feature subsets for the ML models.

Our main contributions are summarized as:
\begin{enumerate}
\item \textbf{Introduction of a Novel Feature Masking Framework:} We introduce a new framework for feature selection that leverages a feature masking approach. This approach enables our framework to operate directly with the original model without requiring retraining. Thus, unlike traditional wrapper feature selection methods that require retraining the model for each subset of features or the embedded methods that require the model to have an inherent feature importance attribute, our approach allows efficient feature selection without the need for retraining for generic ML models. Our approach is also more advantageous compared to the filter methods as the interactions between the features and the model are considered in our framework.
\item \textbf{Development of GBMO and FLBMO Algorithms:} We introduce two algorithms: GBMO and FLBMO. Both of these algorithms are based on our feature masking approach. GBMO is adaptive and optimizes the number of selected features based on the model performance, while FLBMO allows for a fixed number of features, providing flexibility based on application needs.
\item {\textbf{Comprehensive Evaluation:}  We perform extensive experiments on synthetic and widely used real life datasets, demonstrating that our methods consistently outperform traditional feature selection techniques, with up to a 49\% reduction in mean squared error in some cases.}
\item \textbf{Reproducibility:} We publicly share our code for experimental reproducibility and to facilitate further research \footnote{\href{https://github.com/mefe06/feature-selection}{https://github.com/mefe06/feature-selection}}.
\end{enumerate}

The organization of the paper is as follows. We define the feature selection problem in Section \ref{sec:pd}. Then, we provide the introduced solutions in Section \ref{sec:is}. In Section \ref{sec:ex}, we demonstrate the performance of the introduced solutions over an extensive set of experiments on a synthetic dataset and the well-known real-life feature selection datasets. Finally, in Section \ref{sec:conclusion}, we conclude the paper with the final remarks.  

\section{Problem Description}\label{sec:pd}

In this paper, column vectors are represented using the bold lowercase letters, matrices are denoted using the bold uppercase letters, and sets are denoted using the mathematical calligraphic uppercase letters. Vector and matrix entries are real. For a vector $ \bm{x} $ and a matrix $\bm{X}$, their respective transposes are represented as $ \bm{x}^T $ and $ \bm{X}^T $. The symbol $ \odot $ denotes the Hadamard product, the element-wise multiplication operation between two vectors or two matrices. ${x}_{k}$ and ${x}_{k,t}$ represent the $k^{th}$ elements of the respective vectors $\bm{x}$ and $\bm{x}_{t}$, where $t$ is the iteration number. For a matrix $ \bm{X} $, $X_{ij} $ indicates the element in the $ i^{th} $ row and the $j^{th} $ column, $\bm{X}_{(\cdot, \mathcal{J})}$ denotes the submatrix of $\bm{X}$ obtained by selecting all the rows and only the columns whose indices are in the set $\mathcal{J}$, i.e., selecting these features.
The operation $ \sum(\cdot) $ calculates the sum of the elements of a vector or a matrix. Given a vector \( \bm{x} = [x_1, x_2, \ldots, x_n]^T \), the \(L_0\) norm is defined as $\| \mathbf{x} \|_0 = \sum_{i=1}^{n} \mathbb{I}(x_i \neq 0)$, where \( \mathbb{I}( \cdot ) \) is the indicator function, yielding 1 if the condition is true, and 0 otherwise. $|\mathcal{J}|$ denotes the cardinality of the set $\mathcal{J}$.

We study the feature selection problem for both classification and regression tasks. Our goal is to select a subset of features from a dataset to train an ML model and improve its performance on unseen data compared to training on all the features. ML algorithms are prone to overfitting when there are limited examples in the training data to train a model \cite{bishop2006}. This case occurs either when we have many features with not enough examples and strong models or when there is a limited number of examples and a weak model \cite{giraud2015introduction}. For instance, when the feature vector dimensions are comparable to the size of the training data or when the data is non-stationary, there may not be enough samples for the ML model to learn the patterns in the data. To solve this, we introduce a novel feature selection approach, which selects features to avoid overfitting in different data regimes. Our approach has automatic feature selection, and the number of the features can be specified beforehand if needed for different applications.

We define the feature selection problem as follows. Consider a dataset, $\mathcal{D} =  \{(\bm{x}_i, {y}_{i})\}^{N+K}_{i=1}$, where $N$ is the number of samples in the training split, $K$ is the number of samples in the validation split, $\bm{x}_i \in \mathbb{R}^{M}$ is a feature vector and $M$ is the size of a feature vector, and $y_{i}$ is the corresponding target variable. The target variable is a real number in regression and categorical in classification tasks. Given an ML model $f(\cdot ,\cdot)$, the model parameters $\bm{\theta}$, and a feature vector $\bm{x}_i$, the estimate of the model is given by:
\begin{equation}
\hat{y}_{i} = f(\bm{x}_i, \bm{\theta})\text{.}
\end{equation}
We use the same $f(\cdot ,\cdot)$ for feature matrices with different numbers of features to simplify the notation with an abuse of notation. Naturally, these different $f(\cdot ,\cdot)$'s do not have the same number of parameters.

We define $\bm{X}=[\bm{x}_1, \ldots, \bm{x}_N]^T$ as the collection of feature vectors, i.e., each row is the transpose of a feature vector. We define
\begin{equation}
f_N(\bm{X}, \bm{\theta}) = [f(\bm{x}_1, \bm{\theta}), f(\bm{x}_2, \bm{\theta}), \ldots, f(\bm{x}_N, \bm{\theta})]^T 
\end{equation}
as the output vector for this dataset.
For a single estimate, we suffer the loss $l(\hat{y}_{i}, y_{i})$. Further, we define $\bm{y}$ and $\hat{\bm{y}}$ as $[y_1, \ldots, y_N]^T$ and $[\hat{y}_1, \ldots, \hat{y}_N]^T$, respectively, hence, $\hat{\bm{y}}= f_N(\bm{X}, \bm{\theta})$. 
For a given loss function $l(\cdot, \cdot)$, a feature matrix $\bm{X}$, the corresponding target vector $\bm{y}$, and the model parameters $\bm{\theta}$, the model suffers the total normalized loss
\begin{equation}
\mathcal{L}(f_N(\bm{X}, \bm{\theta}), \bm{y}) = \frac{1}{N}\sum_{i=1}^{N}l(\hat{y}_{i}, y_{i}) \text{.}
\end{equation}

In the feature selection problem, we first train the ML model. Then, the feature selection algorithm chooses certain features from the given feature set based on the given criteria. We then train the same ML model on the chosen features. In addition, we have an unseen feature set with the same features. Our goal is to select the features such that the performance of the model trained on the selected features is maximized in this unseen dataset, i.e., optimizing the generalization performance under the given loss. 

We first optimize the given ML model by minimizing the total loss with respect to the set of model parameters $\bm{\theta}$ as
\begin{equation} \label{eq:theta}
\hat{\bm{\theta}} =  \underset{\bm{\theta}}{argmin} \;\:  \mathcal{L}(f_N(\bm{X}, \bm{\theta}), \bm{y} )
\end{equation}
and obtain $\hat{\bm{\theta}}$ as the optimal parameters. Based on this trained model, we next choose a subset of the features using a feature selection algorithm on the validation split. We define $\bm{X}^{'} = \bm{X}_{(\cdot, \mathcal{J})}$ where $\mathcal{J}$ is the set of the indices of the selected features. We again train our ML model on $\bm{X}^{'}$ and $\bm{y}$ as
\begin{equation}
\label{eq: opt}
    \hat{\bm{\theta}}^{'} =  \underset{\bm{\theta}^{'}}{argmin} \;\:  \mathcal{L}(f_K(\bm{X}^{'}, \bm{\theta}^{'}), \bm{y} )
\end{equation}
and obtain $\hat{\bm{\theta}}^{'}$.
Next, we provide an unseen feature matrix $\bm{X}_{new}$ with $M$ samples, and the corresponding target variable vector $\bm{y}_{new}$, and select the features from $X_{new}$ using $\mathcal{J}$ as follows:
\begin{equation}
   \bm{X}_{new}^{'} = \bm{X}_{new, (\cdot, \mathcal{J})} 
\end{equation}
where $\bm{X}_{new}^{'}$ is the selected feature subset matrix.
Our feature selection goal is to find an optimal $\mathcal{J}$ that minimizes 
\begin{equation}
\mathcal{J}^* =  \underset{\mathcal{J}}{argmin} \;\: \mathcal{L}(f_M(\bm{X}_{new}^{'}, \hat{\bm{\theta}}^{'}), \bm{y}_{new}) \text{,}
\end{equation}
i.e., minimizes the model loss on the unseen data. Selecting the informative features for the model increases the performance of the model on the unseen data with respect to the loss by increasing the generality of the model and preventing overfitting \cite{overfit}.

Furthermore, as an additional constraint, the desired number of features, say $\eta$, may be fixed. Thus, we may have the additional constraint that $|\mathcal{J}^{*}|=\eta$. 

To this end, we next present our approaches to find $\mathcal{J}^{*}$ for both the general case and the fixed size case. 

\section{Feature Masking Framework for Feature Selection}\label{sec:is}

Here, we use a mask to operate on the feature vectors for feature selection. We define a binary mask vector, $\bm{m} \in \{0, 1\}^M$, where $M$ is the initial number of features in the dataset. The mask selects features from a feature vector by the Hadamard product. A detailed example is provided in Appendix A.

After the feature selection process ends, we remove the zero columns from $\bm{X} \odot [\bm{m}, \ldots, \bm{m}]^T$ to obtain the selected features, i.e., $\bm{X}^{'}$ with $K$ samples, i.e., $K=2$ in this example. Thus, we have the selected feature indices that depend on the mask vector as $\mathcal{J}= \{ j:{m}_{j} = 1\}$, i.e., the set of the indices of the entries with the value 1 in the mask vector. Hence, $\mathcal{J}= \{1, 3\}$ in this example. Then, our feature selection goal becomes the optimization of the binary feature mask with respect to the defined loss, expressed by 
\begin{equation}
\bm{m}^* =  \underset{\bm{m}}{argmin} \;\: \mathcal{L}(f_P(\bm {X}_{new}^{'}, {\hat{\bm{\theta}}}^{'}), \bm{y}_{new}) 
\end{equation}
where $\bm{X}_{new}$ is an unseen arbitrary feature matrix with $P$ samples, $\bm{X}_{new}^{'}=\bm{X}_{new, (\cdot, \mathcal{J})}$, $\bm{y}_{new}$ is the corresponding target variable vector and ${\hat{\bm{\theta}}}^{'}$ is given by
\begin{equation}
    \hat{\bm{\theta}}^{'} =  \underset{\bm{\theta}^{'}}{argmin} \;\: \mathcal{L}(f_K(\bm {X}^{'}, {\bm{\theta}}^{'}), \bm{y}) \text{,}
\end{equation}
from (\ref{eq:theta}). Here, we note that  $\bm{X}_{new}^{'}$ depends on $\bm{m}$ as the indices of selected features $\mathcal{J}$ depend on $\bm{m}$ as explained in the example.

We next describe two different approaches. 
\subsection{General Binary Mask Optimization (GBMO)}
\label{subsec:sec-gbmo}
We introduce our GBMO framework for the feature selection problem. This algorithm facilitates the general binary mask optimization for an ML model. In the next section, we introduce an extension to the GBMO algorithm when we have an additional constraint on the size of the feature mask. We explain both algorithms and present their pseudocodes.

First, consider the initial dataset, $\mathcal{D} =  \{(\bm{x}_i, {y}_{i})\}^{N+K}_{i=1}$, where we split the initial dataset into the training and the feature selection validation splits randomly,
$\mathcal{D}_{train} =  \{(\bm{x}_{train,i}, {y}_{train, i})\}^N_{i=1}$, and $\mathcal{D}_{val} =  \{(\bm{x}_{val, i}, {y}_{val, i})\}^{N+K}_{i=N+1}$, where $\mathcal{D}$ is the combination of $\mathcal{D}_{train}$ and $\mathcal{D}_{val}$. Here, $\bm{x}_{train, i}$ and $\bm{x}_{val, i} \in \mathbb{R}^M$. We optimize the total loss function $\mathcal{L}(\cdot, \cdot)$ with respect to the model parameters $\bm{\theta}$ on $\mathcal{D}_{train}$, and obtain $\hat{\bm{\theta}}$, the optimal parameters, as:
\begin{equation}\label{eq:theta_2}
\hat{\bm{\theta}} =  \underset{\bm{\theta}}{argmin} \;\:  \mathcal{L}(f_N(\bm{X}_{train}, \bm{\theta}), \bm{y}_{train} )\text{.}
\end{equation}
After we obtain the optimal parameters $\hat{\bm{\theta}}$, we select certain features from the dataset to improve the model performance on the unseen data, $\mathcal{D}_{val}$.

We initialize the feature mask as $\bm{m}_0 \in  \{1\}^M$, i.e., all the features are present. In each iteration, we find the least useful feature included in the dataset. We define the least useful feature as the feature that results in the minimum validation loss when it is not used in the predictions of the model, i.e., zeroed out with the mask. Then, we compare the validation loss at each iteration with the validation loss of the previous iteration. If the current validation loss is smaller than the previous validation loss times some scalar larger than 1, the current least useful feature is not informative for the model. If the absence of a feature does not significantly increase the validation loss compared to the validation loss when it is included, it is not informative for the ML model. Further, we multiply the validation loss by a scalar larger than 1 to bias the model to eliminate more features to increase the generalization. Thus, the iterations continue as we find a feature to eliminate. If the validation loss increases, i.e., all the remaining features are informative, we stop the iterations. After the iterations end, we select features from the dataset according to the final mask vector.

We define $\mathcal{J}_t= \{ j:{m}_{j, t} = 1 \}$, i.e., the set of the indices corresponding to the non-zero entries of the mask vector $\bm{m}_t$ at the iteration $t$, i.e., the mask operator may change after each iteration.
After the initializations, i.e., $\bm{m}_0 \in  \{1\}^M$, the optimization iterations start. 

In each iteration $t$, the optimization algorithm finds the least useful feature as follows. We define $\bm{m}^{(j)}_{t-1}$ as
\begin{equation}
{m}^{(j)}_{i, t-1} = 
\begin{cases} 
0 & \text{if } i = j \\
{m}_{i, t-1} & \text{otherwise,} 
\end{cases} 
\end{equation}
i.e., a copy of $\bm{m}_{t-1}$ where we update the entry at the $j^{th}$ index as 0, where $\bm{m}_{t-1}$ is the mask vector at the ${(t-1)}^{th}$ iteration.
Following the definition of $\bm{m}^{(j)}_{t-1}$, we define $\bm{M}^{(j)}_{t-1}$ as $\bm{M}^{(j)}_{t-1} = \bm{1}{\bm{m}^{(j)}_{t-1}}^T$, where $\bm{1} \in \{1\}^K$, i.e., a matrix obtained by repeating $\bm{m}_{t-1}$ vector $K$ times and setting the $j^{th}$ column of the matrix as zero. Using these definitions, we define the least useful feature index $j_{t}^{*}$ for the iteration $t$ as:
\begin{equation}\label{eq:sluf}
j_{t}^{*} =  \underset{j\in \mathcal{J}_{t-1}}{argmin} \;\:  \mathcal{L}(f_K(\bm{X}_{val}\odot \bm{M}^{(j)}_{t-1},\hat{\bm{\theta}}), \bm{y}_{val}) \text{,}
\end{equation}
i.e., the least useful feature at the iteration $t$ is the feature that results in the minimum validation loss when its value is set to 0 for all the samples out of the included features, where $\hat{\bm{\theta}}$ is defined in (\ref{eq:theta_2}). If the validation loss is small when a feature is not used in the predictions, i.e., zeroed out in our case, then the corresponding feature is less informative for the predictions of the model. 
Using the least useful feature index $j_{t}^{*}$, we obtain the minimum validation loss $Loss_t^{min}$ for the iteration $t$ as:
\begin{equation}
Loss_t^{min} = \mathcal{L}(f_K(\bm{X}_{val}\odot \bm{M}^{(j_{t}^{*})}_{t-1}, \hat{\bm{\theta}}), \bm{y}_{val})\text{.}
\end{equation}
This loss is compared to the loss of the previous iteration. If $Loss_t^{min}<Loss_{t-1}^{min}\times(1+\mu)$, where $\mu$ is the slack hyperparameter, we update the mask $\bm{m}_{t-1}$ by,
\begin{equation}
{m}_{i, t} = 
\begin{cases} 
0 & \text{if } i = j_{t}^{*}  \\
{m}_{i, t-1} & \text{otherwise} 
\end{cases}\end{equation}
to obtain the mask vector $\bm{m}_{t}$, i.e., we zero out the feature with index $j_{t}^{*}$ as the absence of this feature does not have recognizable effect on the performance of the algorithm on the validation dataset. The slack hyperparameter determines the likeliness of the algorithm to eliminate features. It can be determined by validation as shown in our experiments.
If $Loss_t^{min} >Loss^{min}_{t-1}\times(1+\mu)$, we stop the iterations as the validation loss increases significantly, which implies all the remaining features are informative.

We denote the final iteration number as $t^{'}+1$. After the optimization iterations stop at the iteration $t^{'}+1$, we eliminate the features from the training dataset using the final mask vector, $\bm{m}_{t^{'}}$. This corresponds to
\begin{equation}\label{eq:set}
\bm{X}^*_{train} = \bm{X}_{train, (\cdot, \mathcal{J}_{t^{'}})}
\end{equation}
where $\mathcal{J}_{t^{'}}=\{ j:{m}_{j, t^{'}} = 1 \} $, and $\bm{X}^*_{train}$ is the selected feature matrix for the model.
Then, we optimize the model on the updated feature matrix $\bm{X}^*_{train}$ with respect to the loss function to obtain the optimal parameters: 
\begin{equation}\label{eq:new_param}
\hat{\bm{\theta}}^{'}_{GBMO} =  \underset{\bm{\theta}^{'}}{argmin} \;\:  \mathcal{L}(f_N(\bm{X}^*_{train}, \bm{\theta}^{'}), \bm{y}_{train}) \text{,}
\end{equation}
hence we have the optimized model, $f(\cdot, \hat{\bm{\theta}}_{GBMO}^{'})$. Additionally, we use cross-validation and a separate validation set to tune the hyperparameters of the ML model and the feature selection framework, respectively. We explain these steps in depth in Section \ref{ssec:splits}.

The pseudocode for selecting the least useful feature is given in Algorithm \ref{alg:sluf}. The pseudocode for the binary feature mask optimization is given in Algorithm \ref{alg:gbmo}.

\begin{algorithm}
\caption{Select Least Useful Feature (SLUF)}\label{alg:sluf}
\begin{algorithmic}[1]
\State{Inputs $\bm{m}_{t} \in  \{0, 1\}^M$, $\mathcal{D}_{val}, f(\cdot, \bm{\theta}), \mathcal{L}(\cdot, \cdot)$}
\State{$MinLoss \gets \infty$}
\State {$\mathcal{D}_{val}=\{(\bm{x}_i, y_i)\}_{i=N+1}^{N+K}$}
\State {$\bm{x}_i \in \mathbb{R}^M \; \forall i \in \{1,\ldots,K\}$}
\State {$\mathcal{J}_t := \{ j:{m}_{j, t} = 1 \}$}
\State{${m}^{(j)}_{i, t} := 
\begin{cases} 
0 & \text{if } i = j \\
{m}_{i, t} & \text{otherwise} 
\end{cases}$}
\State{{$\bm{M}^{(j)}_{t} = \bm{1}{\bm{m}_{t}^{(j)}}^T, \bm{1}\in \{1\}^K$}}
\State{$j_{t+1}^{*} =  \underset{j\in \mathcal{J}_{t}}{argmin} \;\:  \mathcal{L}(f_K(\bm{X}\odot \bm{M}^{(j)}_{t},\hat{\bm{\theta}}), \bm{y} )$}
\State{$Loss_{t+1}^{min} = \mathcal{L}(f(\bm{X}\odot \bm{M}^{(j_{t+1}^{*})}_{t}, \hat{\bm{\theta}}), \bm{y})$}
\\
\algorithmicreturn { $j_{t+1}^{*} , Loss_{t+1}^{min}$} 
\end{algorithmic}
\end{algorithm}

\begin{algorithm}
\caption{General Binary Mask Optimization}\label{alg:gbmo}
\begin{algorithmic}[1]
\State{Inputs $\mathcal{D}, f(\cdot, \cdot),\mathcal{L}(\cdot, \cdot), \mu$}
\State $\mathcal{D}=\{(\bm{x}_i, y_i)\}_{i=1}^{N+K}$
\State $\bm{x}_i \in \mathbb{R}^M \; \forall i \in \{1,\ldots,N+K\}$
\State $\mathcal{D}_{train}, \mathcal{D}_{val} \gets \mathcal{D}$
\State $\bm{m}_{0} \gets \{1\}^M$ 
\State $\hat{\bm{\theta}} =  \underset{\theta}{argmin} \;\:  \mathcal{L}(f_N(\bm{X}_{train}, \bm{\theta}), \bm{y}_{train})$ 
\State $t= 1$
\State $Converge = False$
\State{$Loss_{0}^{(min)} \gets \infty$}
\While{$not \text{ }Converge$}
    \State{$j_{t}^{*}, Loss_{t}^{(min)} = SLUF(\bm{m}_{t-1},\mathcal{D}_{val}, f(\cdot, \hat{\bm{\theta}}), \mathcal{L}(\cdot, \cdot))$}
    \If{$(Loss_{t}^{min}>Loss_{t-1}^{min}(1+\mu))$}
        \State{$Converge = True$}
    \Else
        \State{${m}_{i, t} = 
        \begin{cases} 
        0 & \text{if } i = j_{t}^{*} \\
        {m}_{i, t-1} & \text{otherwise} 
        \end{cases}$}        
    \EndIf
    \State{$t = t+ 1$}
\EndWhile \\
\algorithmicreturn { $\bm{m}_{t-1}$} 

\end{algorithmic}
\end{algorithm}
In Algorithm \ref{alg:sluf}, we take the mask vector, $\bm{m}_{t}$, the feature selection validation dataset $\mathcal{D}_{val}$, the loss function $\mathcal{L}(\cdot, \cdot)$, and the ML model with parameters $\bm{\theta}$, $f(\cdot, \bm{\theta})$, as inputs. Then, we define the set $\mathcal{J}_t$, the vector $\bm{m}^{(j)}_t$, and the matrix $\bm{M}^{(j)}_t$ in lines 5-7. In line 8, we find the least useful unmasked feature with the index $j^*_{t+1}$. In line 9, we calculate the loss $Loss_{t+1}^{min}$ when the ${j^*_{t+1}}^{th}$ feature is masked. We return $j^*_{t+1}$ and $Loss_{t+1}^{min}$.

In Algorithm \ref{alg:gbmo}, we take the initial dataset $\mathcal{D}$, the ML model $f(\cdot, \cdot)$, the loss function $\mathcal{L}(\cdot, \cdot)$, and the model hyperparameter $\mu$ as inputs. We then split the dataset into the training and the validation splits in line 4. We initialize the mask vector in line 5. Then, the mask optimization iterations start. In line 11, we obtain the least useful feature index $j^*_t$ for the iteration $t$. In line 12, we compare the current minimum loss with the minimum loss of the previous iteration scaled by $1+ \mu$. If the current loss is greater, we change the $Convergence$ value as $True$. Otherwise, we update the mask by updating its $j^{*^{th}}_t$ entry as 0. We iterate over the lines 11-17 and check the stopping condition in line 10 in each iteration. The iterations continue until $Convergence = True $. We then return the final mask vector $\bm{m}_{t-1}$.
\subsection{Binary Mask Optimization For a Determined Number of Features (FLBMO)}
\label{subsec:sec-flbmo}

In this section, we explain FLBMO for the feature selection problem where the number of features is set beforehand, e.g., due to the hardware or computational constraints. In this setting, the desired mask vector has a constraint in the $L_0$ norm.

In this case, we eliminate the least useful feature in each iteration $t$ with the corresponding index $j^*_t$ as in (\ref{eq:sluf}). After eliminating the least useful feature, we check the stopping condition, which is the number of the unmasked features, $\left\| \bm{m}_{t} \right\|_0$. If $\left\| \bm{m}_{t} \right\|_0 =\eta$, then the optimization stops at the iteration $t+1$. Here, $\eta$ is the preset hyperparameter for tuning the number of the unmasked features, which directly determines how many features we keep at the end of the feature elimination iterations. We obtain the final mask $\bm{m}_t$.
After obtaining the mask, we follow (\ref{eq:set}) to select the features according to the mask $\bm{m}_t$ and update the feature matrix as $\bm{X}^*_{train}$. We then use
\begin{equation}
\hat{\bm{\theta}}_{FLBMO}^{'} =  \underset{\bm{\theta}^{'}}{argmin} \;\:  \mathcal{L}(f_N(\bm{X}^*_{train}, \bm{\theta}^{'}), \bm{y}_{train})
\end{equation}
to obtain the optimal parameters $\hat{\bm{\theta}}_{FLBMO}^{'}$ on the updated dataset. We perform hyperparameter tuning for the ML model and the feature selection framework. We explain this in Section \ref{ssec:splits} in detail.
We present the procedure at Algorithm \ref{alg:flbmo}.
\begin{algorithm}
\caption{Fixed Length Binary Mask Optimization (FLBMO)}\label{alg:flbmo}
\begin{algorithmic}[1]
\State{Inputs $\mathcal{D}, f(\cdot, \cdot), \mathcal{L}(\cdot, \cdot), \eta$}
\State $\mathcal{D}={(\bm{x}_i, y_i)}_{i=1}^{N+K}$
\State $\bm{x}_i \in \mathbb{R}^M \; \forall i \in \{1,\ldots,N+K\}$
\State $\mathcal{D}_{train}, \mathcal{D}_{val} \gets \mathcal{D}$
\State $\bm{m}_{0} \gets \{1\}^M$ 
\State $\hat{\bm{\theta}} =  \underset{\theta}{argmin} \;\:  \mathcal{L}(f_N(\bm{X}_{train}, \bm{\theta}),\bm{y}_{train})$ 
\State $t= 1$
\While{$\| \bm{m}_{t-1} \|_0 >\eta $}
    \State{$j_{t}^{*}, Loss_{t}^{min} = SLUF(\bm{m}_{t-1},\mathcal{D}_{val}, f(\cdot, \hat{\bm{\theta}}), \mathcal{L}(\cdot, \cdot))$}
    \State{${m}_{i, t} = 
    \begin{cases} 
    0 & \text{if } i = j_{t}^{*} \\
    {m}_{i, t-1} & \text{otherwise} 
    \end{cases}$}
    \State{$t = t+ 1$}

\EndWhile \\
\algorithmicreturn { $\bm{m}_{t-1}$} 

\end{algorithmic}
\end{algorithm}

In Algorithm 3, we take the initial dataset $\mathcal{D}$, the ML model $f(\cdot, \cdot)$, the loss function $\mathcal{L}(\cdot, \cdot)$, and the model hyperparameter $\eta$ as inputs. We split the dataset into the training and the validation datasets in line 4. We initialize the mask vector in line 5. Then, the mask optimization iterations start. In line 9, we select the least useful feature with the index $j^*_t$ for the iteration $t$. In line 10, we update the mask by setting its $j^{*^{th}}_t$ entry as 0. We iterate over the lines 9-11 until $\left\| \bm{m}_{t} \right\|_0=\eta$. We then return the final mask vector $\bm{m}_t$.

\section{Experiments}\label{sec:ex}

In this section, we evaluate the performance of our feature selection models, GBMO and FLBMO. First, we conduct experiments on a synthetic dataset where we know which features are informative, i.e., under controlled settings. Then, for performance comparisons, we use the well-known real-life benchmark datasets. We include both regression and classification tasks with different models to demonstrate the performance of our models. We select our loss functions as the mean squared error (MSE) and the log loss for regression and classification tasks, respectively, for our experiments. However, as emphasized, our approach can be used for the other loss functions. 
\subsection{Datasets}\label{ssec:datasets}
We first use a synthetic dataset to evaluate the feature selection performance of our algorithms. Here, we have a complex non-linear relation between the informative features and the target variable. We demonstrate that our algorithm chooses the informative features out of all the features, which also include the redundant features. Next, we use two real-life datasets that contain a high number of features compared to the number of samples to demonstrate the effectiveness of our algorithms.
All the real-life datasets are obtained from the UCI Machine Learning Repository \cite{dataset}. Next, we present the short descriptions of the datasets.
\subsubsection{Connectionist Bench}
The goal of this dataset is to predict whether the sonar signals bounced off a metal cylinder or a rock \cite{misc_connectionist}. We have a binary classification dataset with 60 features and 208 samples. The features represent the measured energies in the different frequency bands. 
\subsubsection{Residential Building}
The goal of this dataset is to predict the construction costs and the sale prices of the residential apartments in Iran \cite{misc_residential_building_data_set_437}. We only predict the construction costs in our experiments for simplicity. This is a regression dataset with 103 features and 372 samples. The features include the variables about the project finances and the general economic information.

Each dataset is split into 45\% training that corresponds to $\mathcal{D}_{train}$ in (\ref{eq:theta_2}), 30\% feature selection validation for GBMO and FLBMO that corresponds to $\mathcal{D}_{val}$ in (\ref{eq:sluf}), 10\% feature selection model validation, and 15\% test. 

\subsection{Models and Hyperparameter Tuning}\label{ssec:splits}
We use LightGBM and MLP as the ML models in the experiments. We prefer these models because of their demonstrated performance in the well-known data competitions \cite{ke2017, dl}. As the comparison feature selection algorithms, we use the RFE feature selection \cite{guyon2003}, the MI feature selection \cite{vergara2014review}, the CC feature selection \cite{farahani2020feature}, which are well-known, and our algorithms, GBMO and FLBMO. Further, we include the model performance using all the features as a baseline. 
We select the best set of hyperparameters based on the MSE and log loss on the validation sets for regression and classification tasks, respectively. Next, we explain how the we use the split data in order to obtain an optimized ML model on the selected features.

Initially, we randomly split the dataset to three parts: 75\% for ML model training, 10\% for feature selection model validation, 15\% for testing. We need to split the dataset to four parts for the training and evaluation of our algorithms, GBMO and FLBMO: ML model training, feature selection validation, feature selection model validation, and test. We use the initial split and randomly split the ML model training subset to ML model training and feature selection validation subsets to obtain 4 subsets from the original dataset. Here, we split the initial ML model training subset so that the new ML model training subset and feature selection validation subset correspond to 45\% and 30\% of the original dataset, respectively. We first get the best hyperparameters for the ML model with cross-validation and train the ML model with these hyperparameters on the training split. Then, our algorithms need a separate set for feature selection, which is the feature selection validation set. Then, we use another separate set to validate the hyperparameters of the feature selection algorithms. With this process, we obtain the best feature selection framework with the best hyperparameters selecting the best features for the best ML model for the dataset. Finally, we evaluate the test performance of the obtained ML model on the test split.

For all feature selection algorithms, we used the same samples to ensure a fair comparison of model performance. For traditional feature selection approaches, we need three splits: ML model training, feature selection model validation, and test. Here, we use the initial split with 3 parts. Thus, the ML model training subset is the union of ML model training and feature selection validation subsets used in our approaches. The other subsets, i.e., the model validation and the test, are the same in our introduced approaches and the other feature selection algorithms, ensuring that all models were evaluated on identical samples. For the other algorithms, we use the ML model training split for both obtaining the best ML model with cross-validation and also selecting the features. We again validate the feature selection approaches on the feature selection model validation set. Then, we obtain the test score for each feature selection approach on each ML model using the test set. The hyperparameter search space is provided in Table \ref{tab:hyperparameters}. Here, for the feature selection methods that require a fixed number of features, we select the search space for the hyperparameter $\eta$ according to the number of features $M$ in the dataset.

\begin{table}[ht]
\centering
\caption{Hyperparameter Search Space}
\label{tab:hyperparameters}
\begin{tabular}{ll}
\hline
\textbf{Model Name} & \textbf{Search Space} \\
\hline
MLP     & hidden\_layer\_sizes: \{(20),(40),(10),(20,10)\}, \\
        & activation: \{'relu', 'logistic'\}, \\
        & alpha: \{0.0001, 0.001, 0.01\}, \\
        & learning\_rate\_init: \{0.001, 0.01\} \\
\hline
LightGBM & num\_leaves: \{7, 15\}, \\
         & learning\_rate: \{0.01, 0.025, 0.05\}, \\
         & n\_estimators: \{10, 20\}, \\
         & subsample: \{0.6, 0.8\}, \\
         & colsample\_bytree: \{0.6, 0.8\}, \\
         & min\_child\_samples: \{5, 10\} \\
\hline
GBMO    & $\mu$: \{0.00025, 0.001, 0.01, 0.05\} \\
\hline
FLBMO   & $\eta$: \{M/6,M/5,M/4,M/2\} \\
\hline
CC      & $\eta$: \{M/6,M/5,M/4,M/2\} \\
\hline
MI      & $\eta$: \{M/6,M/5,M/4,M/2\} \\
\hline
RFE     & $\eta$: \{M/6,M/5,M/4,M/2\} \\
\hline
\end{tabular}
\end{table}

\subsection{Synthetic Data Experiments}
 We conduct our first experiment with our feature selection methods on a synthetic dataset to verify the feature selection abilities of our models. The synthetic feature matrix $\bm{X}$ contains 300 samples and 100 features, i.e., $M$= 100. The target variable vector $\bm{y}$ is obtained by
\begin{equation}
y_i = \sum_{j=0}^{9} X_{ij}^2 + sin(X_{ij}) \text{,}
\end{equation}
i.e., only the first 10 features are informative for each sample. 
We split the dataset as explained in Section 4.1. Then, we use the traditional feature selection methods and our approaches to select features from this dataset. Here, our ML model is LightGBM. We have $\{6, 10, 15, 20\}$ as the hyperparameter search space of $\eta$ for FLBMO, RFE, MI, and CC. The experiment results are displayed in Table \ref{tab:synth_results}. 

The ability of GBMO to converge to exactly 10 features without requiring a predetermined number of features to select shows its adaptive nature. This adaptability is important in real-world applications where the exact number of informative features is often unknown. Traditional methods like CC and MI, while useful, exhibit clear limitations in this context. CC's inability to capture non-linear dependencies is a known drawback, and our results affirm this. MI, although capable of identifying non-linear relationships, does not take into account how the model uses the features, which can lead to a disconnect between the selected features and which features are important to the model.

RFE, which relies on the feature importance attribute of the model, performs better but still falls short compared to our methods. The holistic approach of GBMO and FLBMO, which uses the predictions of the model to iteratively refine feature selection, provides a more effective solution. Our methods consider not just the individual feature importances but the combined effect of feature subsets, which is an advantage in complex datasets.

Moreover, the fact that our methods did not select any redundant features demonstrates their precision. Our results show that GBMO and FLBMO are not only capable of identifying the most informative features but also do so in a way that enhances the overall efficiency.

\begin{figure}[ht]
    \centering
    \includegraphics[width=0.75\linewidth]{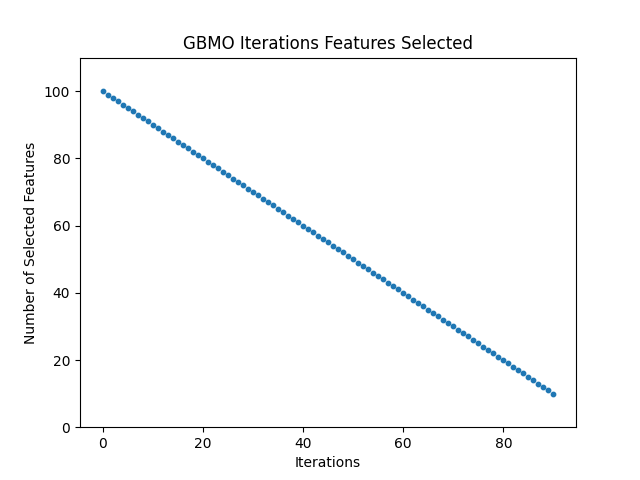}
    \caption{The plot of the number of the remaining features for the GBMO algorithm with respect to the iteration numbers on the synthetic dataset with LightGBM model.}
    \label{fig:fts_iters_synth}
\end{figure}

Further, we investigate the convergence of the GBMO algorithm on the synthetic dataset with LightGBM as the ML model. First, from Fig. \ref{fig:fts_iters_synth}, we see that a feature is eliminated in each iteration, and the iterations stop the first time we find a feature whose absence increases the validation loss above the scalar threshold i.e., $1+\mu$. In addition, we present the validation losses with respect to the iteration numbers in Fig. \ref{fig:val_loss_synth}. In the figure, we see that initially the validation loss decreases when approximately the first 25 features eliminated. Then, the validation loss does not change through 40 iterations as LightGBM does not use these features in constructing the trees. In the last 25 iterations, the eliminated features increase the validation loss slightly. However, we note that even though the absence of these features increases the validation loss slightly, we know these are redundant features as we control the experiment settings. This shows that the significance of the introduced slack hyperparameter, $\mu$, in feature selection, as the last 25 features would not be eliminated if we did not include a slack to increase the inclination of the framework to eliminate features.

\begin{table}[ht]
    \caption{The number of informative and redundant features selected by each feature selection method on synthetic data.}
    \centering
    \label{tab:synth_results}

    \begin{tabular}{ccc}
            \toprule

         Method & Informative Features  & Redundant Features \\
\midrule
         GBMO&10  &0 \\
         FLBMO&10  &0 \\
         CC&7  &13 \\
         MI&10  &5 \\
         RFE&10  &5 \\
\bottomrule
    \end{tabular}

    \label{tab:my_label}
\end{table}
\subsection{Real Life Data Experiments}
As RFE requires the model to have an intrinsic feature importance attribute, the RFE feature selection is not used for the MLP model. Our experiment results for MLP and LightGBM are displayed in Table \ref{tab:MLP_results} and Table \ref{tab:LGBM_results}, respectively. In both real-life datasets, our algorithms surpass the performance of the other methods. Further, we present the number of the selected features by the feature selection methods for the different datasets in Table \ref{tab:f_numbers_1} and Table \ref{tab:f_numbers_2}.

We note that the superior performance of the GBMO algorithm is evident from the experiments. The distinct advantage of GBMO is its adaptive approach that enables it to select the informative features without any constraints on the number of features it selects. This enables GBMO to select different numbers of features for the different datasets as it can be seen from Table \ref{tab:f_numbers_1} and Table \ref{tab:f_numbers_2}. This allows us to select features without needing information about how many features to select for a specific dataset beforehand. In contrast, traditional methods require the number of features to be set as a hyperparameter, posing a challenge when this information is not readily available. While expanding the hyperparameter search space could address this, it is computationally expensive.

FLBMO, while requiring a predefined feature count like RFE, consistently outperforms RFE, as seen in our experiments. This suggests that the methodology of FLBMO of integrating the predictions of the model during feature selection provides a more accurate reflection of which features truly contribute to model performance. The better test scores achieved by FLBMO are indicative of its enhanced ability to identify features that improve generalization to unseen data.

These results highlight the strengths of both GBMO and FLBMO in real-world applications. GBMO’s ability to adapt to the data without needing the number of features to select hyperparameter makes it suitable for real datasets where the number of informative features are not known. FLBMO’s superiority over RFE suggests that in cases where a fixed number of features is necessary, our approach provides a more refined and effective selection process, resulting in better overall model performance.

\begin{figure}[ht]
    \centering
    \includegraphics[width=0.75\linewidth]{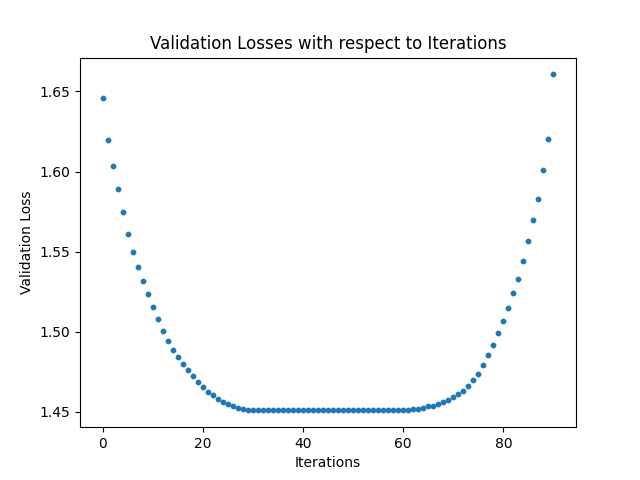}
    \caption{The plot of the validation losses for the GBMO algorithm with respect to the iteration numbers on the synthetic dataset with LightGBM model.}
    \label{fig:val_loss_synth}
\end{figure}

\begin{figure}[ht]
    \centering
    \includegraphics[width=0.75\linewidth]{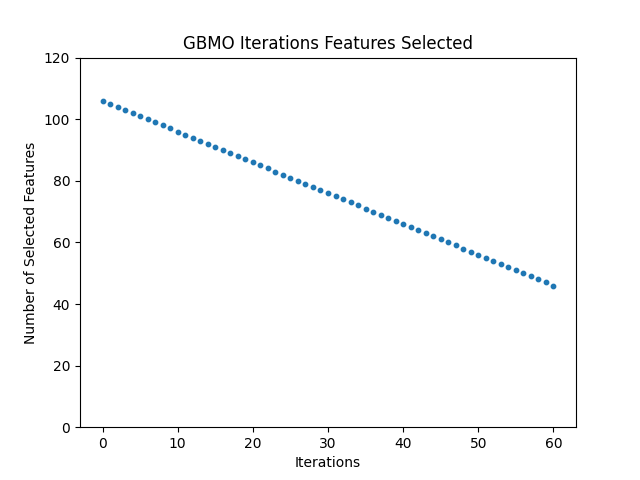}
    \caption{The plot of the number of remaining features for the GBMO algorithm with respect to the iteration numbers on the Residential Building dataset with MLP model.}
    \label{fig:fts_iters_res}
\end{figure}

\begin{figure}[ht]
    \centering
    \includegraphics[width=0.75\linewidth]{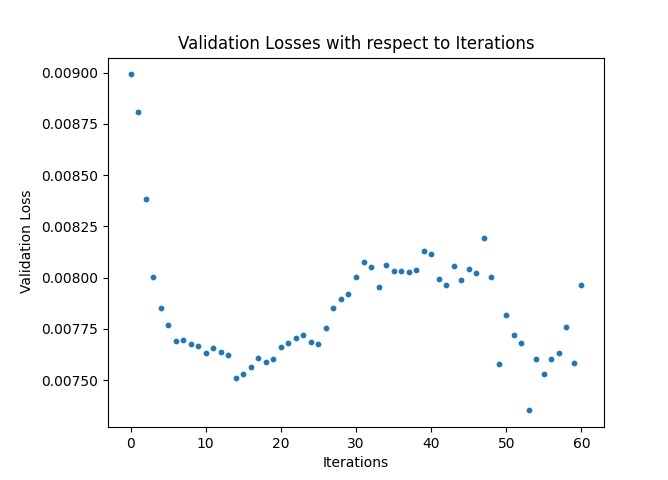}
    \caption{The plot of the validation losses for the GBMO algorithm with respect to the iteration numbers on the Residential Building dataset with MLP model.}
    \label{fig:val_loss_res}
\end{figure}

Further, we investigate the convergence of the GBMO algorithm on the Residential Building dataset with MLP as the ML model. First, from Fig. \ref{fig:fts_iters_res}, we see that a feature is eliminated in each iteration, and the iterations stop the first time we find a feature whose absence increases the validation loss above the scalar threshold, i.e., $1+ \mu$. In addition, we demonstrate the validation losses with respect to the iteration numbers in Fig. \ref{fig:val_loss_res}. We see that the validation loss plot for the MLP model has different characteristics compared to the validation loss plot of the LightGBM model in Fig. \ref{fig:val_loss_synth}. As the LightGBM model does not use some of the features in splitting the trees, the exclusion of some features does not change the validation loss, resulting in a hill shaped graph. On the other hand, MLP has a more irregular graph as MLP gives each feature certain weights. Nonetheless, we see that the GBMO algorithm decreases the validation loss overall and obtains higher validation performance with fewer features.

\begin{table}[ht]
    \caption{The results of the simulations based on the real-life datasets when the ML model is MLP.}
    \vspace*{-5mm}
    \label{tab:MLP_results}
    \centering
        \begin{tabular}{cccccccccc}
            \toprule
            Data  & \multicolumn{1}{c}{Residential Building} & \multicolumn{1}{c}{Connectionist Bench} \\
            Model$\backslash$Metric & MSE($10^{-4}$) & Log Loss \\
            \midrule
            All features &63.18 & 0.1677  \\
            MI &  27.55 & 0.2097  \\
            CC & 20.82 &0.1795   \\
            GBMO & \textbf{18.76} & \textbf{0.1411}  \\
            FLBMO &  37.53 & 0.1466  \\
            \bottomrule
        \end{tabular}
    The best results for each test dataset are shown in bold. 
\end{table}

\begin{table}[ht]
    \caption{The results of the simulations based on the real-life datasets when the ML model is LightGBM.}
    \vspace*{-5mm}
    \label{tab:LGBM_results}
    \centering
        \begin{tabular}{cccccccccc}
            \toprule
            Data & \multicolumn{1}{c}{Residential Building} & \multicolumn{1}{c}{Connectionist Bench} \\
            Model$\backslash$Metric & MSE($10^{-4}$) & Log Loss \\
            \midrule
            All features  & 65.31 & 0.4563  \\
            MI& 86.65 & 0.4957  \\
            CC  & 49.58 & 0.4687  \\
            RFE & 85.30 & 0.4944  \\
            GBMO  & \textbf{44.48} & \textbf{0.3974}  \\
            FLBMO & 45.93 & 0.4301  \\
            \bottomrule
        \end{tabular}
    The best results for each test dataset are shown in bold. 
\end{table}

\begin{table}[ht]
    \caption{The number of the features selected by the feature selection methods in the simulations based on the real-life datasets when the ML model is MLP.}
    \vspace*{-5mm}
    \label{tab:f_numbers_1}
    \centering
        \begin{tabular}{cccccccccc}
            \toprule
            Model$\backslash$Dataset &  \multicolumn{1}{c}{Residential Building} & \multicolumn{1}{c}{Connectionist Bench} \\
            \midrule
            All features  & 105 & 60  \\
            MI &24& 10  \\
            CC &30& 12    \\
            GBMO &46& 45   \\
            FLBMO &20&10  \\
            \bottomrule
        \end{tabular}
\end{table}
\begin{table}[ht]
    \caption{The number of the features selected by the feature selection methods in the simulations based on the real-life datasets when the ML model is LightGBM.}
    \vspace*{-5mm}
    \label{tab:f_numbers_2}
    \centering
        \begin{tabular}{cccccccccc}
            \toprule
            Model$\backslash$Dataset & \multicolumn{1}{c}{Residential Building} & \multicolumn{1}{c}{Connectionist Bench} \\
            \midrule
            All features & 105 & 60  \\
            MI &30& 10   \\
            CC &30& 30   \\
            RFE &60& 12   \\
            GBMO &12& 16   \\
            FLBMO &24& 30  \\
            \bottomrule
        \end{tabular}
\end{table}


\subsection{{Discussion}}

The results of our study highlight the effectiveness of our introduced methods GBMO and FLBMO in feature selection. These methods consistently outperformed traditional approaches, such as MI, CC, and RFE, across diverse datasets, showing the potential of our framework in feature selection.

The primary contribution of our work lies in the development of a novel feature selection framework that eliminates the need for model retraining, a process that is both computationally expensive and impractical in many real-world scenarios. Our approach uses a feature masking technique that works directly with the original model, thus allowing for an efficient selection of features without requiring model training.

The superior performance of our methods, evidenced by reduced test losses compared to traditional approaches, aligns with prior research highlighting the importance of selecting feature subsets that enhance generalization performance \cite{overfit}. Our framework’s ability to dynamically adjust the number of selected features based on model performance without prior knowledge further distinguishes it from traditional approaches, making it highly adaptable to various machine learning tasks.

However, a potential limitation compared to existing literature is our framework's current inability to efficiently handle datasets with a very high number of features. Some methods in the literature adopt a two-phase feature selection approach for high-dimensional datasets \cite{yin2023igrf, xia2023model}, where an initial computationally lightweight feature selection method narrows down the features before applying a more computationally expensive selection method in the second phase. In future work, our approach can be extended to incorporate such hybrid frameworks, enhancing its applicability to high-dimensional datasets.

\subsection{Methodology Limitations}
While our proposed binary feature masking approach offers several advantages, we recognize that there are certain limitations inherent in this methodology. The approach relies on the assumption that the outcomes of the model alone can effectively guide feature selection. However, if the model is not adequately trained or fails to capture the underlying relationships in the data, the performance of our feature selection method may be suboptimal. Additionally, for datasets with a very large number of features, our method may require considerable computational time to identify the final subset of selected features.

\section{Conclusion}\label{sec:conclusion}
We studied feature selection for generic ML models on both regression and classification tasks. We solved the training-free model-specific feature selection problem for generic ML models. To this end, we introduced two algorithms to find the optimal feature subsets that minimize the error on the unseen data both under unconstrained conditions and with a fixed number of features. These algorithms leverage a novel feature masking approach, which eliminates redundant features without requiring retraining of the ML model during feature selection. This novel approach allows us to use a wrapper feature selection approach without training the ML model during the feature selection. Further, we use the predictions of the ML model to select the features, which allows us to use the feature selection framework for generic ML models and also capture the importance of the feature subsets in the predictions of the model. Thus, we obtain a high performance, training-free, model-specific feature selection framework that works on generic ML models. Through extensive experiments on synthetic and well-known real-life datasets with a lower number of samples and a higher number of features, we demonstrate significant performance improvements using our feature selection algorithms compared to the well-established feature selection algorithms in the literature for feature selection. We also openly share the code for our methods and experiments to encourage further research and the reproducibility of our results.

Future research directions could expand the scope and applicability of our methods. Firstly, extending our approach to handle high-dimensional datasets would be a valuable enhancement. This could involve a hybrid feature selection framework, where initial feature filtering is performed using computationally lightweight methods or modified versions of our algorithms, followed by GBMO for improving feature selection to achieve optimal performance. Furthermore, applying our framework to domain-specific real-world applications, such as bioinformatics or finance, could validate its practical utility and inspire tailored adaptations for these fields.

\backmatter
\section*{Declarations}
\begin{itemize}
    \item Data Availability: The data that support the findings of this study is openly available in UCI Machine Learning Repository at \url{https://archive.ics.uci.edu}.
    \item Competing Interests: The authors declare that they have no known
    competing financial interests or personal relationships that could have
    influenced the work reported in this paper.
    \item Ethical Compliance: All the data used in this article are sourced from open and
    publicly accessible platforms. No proprietary, confidential, or private
    data has been used.
    \item Author Contributions: Mehmet E. Lorasdağı: Conceptualization, Methodology, Software, Writing- original draft \& revised manuscript. Mehmet Y. Turalı: Conceptualization, Software, Writing- original draft. Suleyman S. Kozat: Conceptualization, Writing- review \& editing.
    \end{itemize}
\clearpage
\appendix

\section{Appendix: Feature Masking Example}

As an example, given a mask vector $\bm{m}=[1, 0, 1]^T$ and a feature matrix $\bm{X}=[[X_{1,1}, X_{1,2}, X_{1,3}]^T, [X_{2,1}, X_{2,2}, X_{2,3}]^T]^T$, the mask selects features by the Hadamard product, 
\begin{align}
\bm{X} \odot [\bm{m}, \bm{m}]^T &= \begin{bmatrix}
X_{1,1} & 0 & X_{1,3} \\
X_{2,1} & 0 & X_{2,3}
\end{bmatrix}
 \text{,}
\end{align}
i.e., the $2^{nd}$ feature is zeroed out.

\clearpage
\bibliography{sn-bibliography}
\end{document}